\documentclass{article}

\PassOptionsToPackage{compress,numbers}{natbib}

\usepackage[preprint]{neurips_2021}

\usepackage[ref]{leaf}

\usepackage[T1]{fontenc}

\title{A Simple and Fast Baseline for Tuning Large XGBoost Models}

\author{%
  Sanyam Kapoor \thanks{Work done during an internship at Amazon.} \\
  New York University \\
  \texttt{sanyam@nyu.edu} \\
  \And
  Valerio Perrone \\
  Amazon \\
  \texttt{vperrone@amazon.de} \\
}

\begin{document}

\maketitle

\begin{abstract}
XGBoost, a scalable tree boosting algorithm, has proven effective for 
many prediction tasks of practical interest, especially using tabular 
datasets. Hyperparameter tuning can further improve the 
predictive performance, but unlike neural networks, full-batch training of many models on large datasets can 
be time consuming. Owing to the discovery that (i) there is a strong 
linear relation between dataset size \& training time, (ii) XGBoost 
models satisfy the \emph{ranking hypothesis}, and (iii) lower-fidelity 
models can discover promising hyperparameter configurations, we 
show that uniform subsampling makes for a simple yet fast baseline to 
speed up the tuning of large XGBoost models using multi-fidelity 
hyperparameter optimization with data subsets as the fidelity 
dimension. We demonstrate the effectiveness of this baseline on 
large-scale tabular datasets ranging from $15-70\mathrm{GB}$ in size.
\end{abstract}

\section{Introduction}

Despite modern developments in deep learning models for tabular
datasets \citep{Kadra2021RegularizationIA,Somepalli2021SAINTIN}, 
XGBoost \citep{Chen2016XGBoostAS} has stood the test time of time and 
remains the favorite scalable tree boosting algorithm for a wide range 
of problems \citep{ShwartzZiv2021TabularDD}, including large-scale
tabular datasets. Further performance gains can be realized by careful 
hyperparameter optimization (HPO) of XGBoost models.

One of the most successful HPO techniques is sequential Bayesian optimization (BO) \citep{7352306}. BO has consistently 
proven to be the superior method for tuning black-box functions, as was also recently demonstrated by the 
NeurIPS 2020 Black-Box Optimization Challenge 
\citep{Turner2020BayesianOI}. Its sequential nature is, however, 
limiting. XGBoost models with large-scale tabular datasets (i.e. greater than $10\mathrm{GB}$ in size which are our 
focus in this work) come with significant computational 
costs --- training a single model can be time consuming, and 
the full dataset may not even fit the memory. In principle, such models do not allow
stochastic mini-batching as with neural network training.

In this work, however, we establish a surprising result that uniformly subsampling large-scale 
tabular datasets provides a simple, fast, and effective baseline for 
multi-fidelity hyperparameter optimization of XGBoost models. In
particular, we show that:
\begin{itemize}

\item There is a strong linear relationship between the training time of XGBoost models and dataset size (in terms of the fraction of the full dataset). Naturally, training on smaller subsets provides substantial runtime gains.
	
\item Hyperparameter configurations ranked by performance on lower-fidelity versions of XGBoost models (where the fidelity parameter is the fraction of the full dataset size) tend to maintain their relative ranking when trained on the full dataset. This hints that XGBoost models also satisfy the \emph{ranking hypothesis} as discussed for neural network models in \citet{Bornschein2020SmallDB}.
	
\item Tuning lower-fidelity approximations of XGBoost models, with uniformly subsampling as little as $1\%$ of the samples in the full dataset, leads to modest performance drops (often less than $0.5\%$) in terms of the validation score (e.g. AUROC) when compared to well-tuned models on the full dataset.
	
\item For XGBoost models with large-scale tabular datasets, we demonstrate that Hyperband \citep{Li2017HyperbandAN} is much more economical than an exhaustive randomized grid search in terms of the total wallclock time to achieve the same performance. Combining it with BO \citep{Falkner2018BOHBRA} allows us to squeeze out a few more runtime gains.

\end{itemize}

\section{Motivations \& Related Work}

Our main inspiration comes from \citet{Bornschein2020SmallDB}, which 
provides a detailed study of generalization performance of neural 
networks w.r.t dataset size, which complements existing studies w.r.t 
model size. The authors propose the \emph{ranking hypothesis}: 
over-parameterized neural network models tend to maintain 
their relative ranking over a wide range of data subsets drawn from 
the same underlying data distribution.

Our focus, however, is the training of batch models like XGBoost 
\citep{Chen2016XGBoostAS} with very large datasets (often larger than 
$10 \mathrm{GB}$). Neural networks already have the luxury of 
stochastic optimization using mini-batches of data, but XGBoost 
carries a few qualitative differences as it uses all data at once. It 
relies on boosting, i.e., greedily building an additive model by 
adding one base function at a time that learns only the residual 
predictive function. The number of boosting rounds can increase the 
model capacity. This is unlike neural networks, where the model 
capacity is fixed for a given architecture. Further, data (rows) and 
feature (columns) subsampling is already supported by XGBoost, but is 
not to be confused with our goals. XGBoost subsamples for the 
robustness of the constructed ensemble, whereas we are aiming for a 
simple approach to reduce the computational burden of tuning large 
XGBoost models without significantly compromising performance via 
lower fidelity approximations based on data subsets. Notably,
\citet{He2014PracticalLF} briefly describe the use of data subsampling 
in XGBoost models used as feature extractors for logistic regression,
but do not fully explore the computational and performance benefits 
for tuning of XGBoost models.

A large fraction of the literature has focused their analysis on
tuning large neural networks models with stochastic training using
subsets of data \citep{Hartmanis2002NeuralNT,
Bottou2012StochasticGD,Nickson2014AutomatedML}. 
Most recently, \citet{Klein2017FastBO} propose FABOLAS, a general framework to model the loss and training time as a function of the 
dataset size, inspired by multi-task BO 
\citep{Swersky2013MultiTaskBO} where the tasks are now continuous.
The evaluation, however, is still focused on neural network models.
Most notably, the analysis highlights the importance of
using wallclock times for comparing HPO algorithms for practical
usage, which we also do in this work.

With the success of \citet{Bornschein2020SmallDB}, one may be tempted
to believe that the inductive biases of neural networks are aligned
with natural data like images, which form the bulk of the benchmarks,
and are therefore amenable to training using subsets. Tabular 
datasets, however, are not expected to have such easily exploitable 
biases, as has been shown by previous work 
\citep{ShwartzZiv2021TabularDD,Kadra2021RegularizationIA,
Somepalli2021SAINTIN}. Surprisingly, to the contrary, we empirically
demonstrate that batch trained models like XGBoost are also amenable 
to training with uniformly sampled subsets of large datasets.

\section{Background}

There are two broad approaches to achieve scalable and efficient 
hyperparameter optimization: (i) modeling the landscape of 
hyperparameters to model's performance, we can be more efficient about 
\emph{configuration selection}, e.g., through BO 
\citep{7352306}; and (ii) adaptively allocating computational
resources, we can evaluate a large number of hyperparameters, and 
be more efficient about \emph{configuration evaluation}, e.g., through Hyperband \citep{Li2017HyperbandAN}. Both approaches can also be 
combined for further practical gains \citep{Falkner2018BOHBRA}.

\paragraph{Bayesian Optimization (BO)} We can formulate the 
performance of any machine learning model as a function 
$f: \mathcal{X} \to \reals$, where $\mathcal{X}$ is the search space 
of hyperparameter configurations. The hyperparameter optimization 
(HPO) problem can then be defined as the search for the optimal 
configuration $x_\star \in \mathcal{X}$, where 
$x_\star = \argmax_{x \in \mathcal{X}} f(x)$ gives us the globally 
optimal score value (e.g., validation accuracy for classification 
models). BO models this function using a
probabilistic model, $p(f\mid \dset)$, conditioned on the evaluations 
$\dset = \{(x_0, f(x_0)), \dots\}$ observed so far 
\citep{Frazier2018ATO}. We can use this model to query a new 
configuration $x^\prime$ that maximizes an \emph{acquisition function} 
$a(x)$ of interest, i.e., $x^\prime = \argmax_{x\in \mathcal{X}}a(x)$. 
A common choice of the probabilistic model is \emph{Gaussian 
processes} \citep{Rasmussen2005GaussianPF}, and the acquisition 
function is the \emph{Expected Improvement} (EI) 
$a(x) = \mathbb{E}_{p(f\mid\dset)}[\max(0, f(x) - f(x^\star))]$ 
\citep{Mockus1977OnBM}, where $x^\star$ is the best configuration 
seen so far. For subsequent trials, we refit the model with $\dset 
\leftarrow \dset \cup \{(x^\prime, f(x^\prime))\}$, and repeat the
acquisition step.

\paragraph{Multi-Fidelity Hyperparameter Optimization (HPO)} For a given black box function $f(x)$, 
multi-fidelity optimization aims to learn from an augmented function 
$f(x,r)$ with a fidelity parameter $r \in [r_{\min}, r_{\max}]$, such 
that $f(x) = f(x, r_{\max})$ \citep{Swersky2013MultiTaskBO}. In the 
case of HPO, $r$ represents the computational resource. The 
expectation is that low fidelity approximations of the true function, 
i.e., $r < r_{\max}$, are computationally much cheaper, but informative 
towards learning $f(x)$. Popular choices of the fidelity parameter $r$ 
include the number of epochs when training neural networks, or the 
fraction of the full dataset used for training the model.

\paragraph{Hyperband (HB)} Framing the hyperparameter optimization 
as a multi-armed bandit problem, Hyperband \citep{Li2017HyperbandAN}
is an approach towards multi-fidelity HPO that builds upon repeated 
trials of \emph{Successive Halving} (SH)
\citep{Jamieson2016NonstochasticBA}. For a total computational budget 
$B$, SH uses the average budget $B/r$ for each hyperparameter 
configuration $B/r$, where $r$ is fixed a priori. This leads to a 
trade off --- a small value of $r$ would allow many evaluations, but 
at lower and less reliable fidelities, whereas a large value of $r$ 
would allow only a small number of reliable evaluations. Hyperband 
instead uses multiple trials of SH (``\emph{brackets}") for different 
values of $r$ (``\emph{rung levels}"). At each bracket, Hyperband 
ranks the different configurations, only allowing the top $1/\eta$ 
fraction to continue to a higher rung level. Hyperband relies on
random draws of the hyperparameter configurations for convergence to
the global optimum, and often works very well for small to medium
computational budget. By accounting for information from existing 
evaluations, BOHB \citep{Falkner2018BOHBRA} improves upon 
Hyperband by combining BO with HB.

\section{Experiments}

\paragraph{Datasets} In our benchmark study, we focus on large-scale 
tabular datasets which are at least $10$ GB in raw size. The actual
size after feature preprocessing is often much larger. We keep the
feature preprocessing to a minimum as provided by AWS Sagemaker 
\citep{PDas2020, Perrone2021AmazonSA}, which includes converting text
features into \textrm{tf-idf} vectorization 
\citep{BaezaYates2011ModernIR}, categorical variables into one-hot 
representation, and splitting datetime variables across 
days/weeks/months.\footnote{The complete set of feature processors and their implementation is available at \url{https://github.com/aws/sagemaker-scikit-learn-extension}}  We include datasets that are both 
classification and regression to demonstrate the generality of our 
results. We use a uniform random 80/10/10 split for train/validation/test. The complete list of benchmark datasets and their key 
details are provided in \cref{tab:datasets}. All the used datasets are publicly available at \texttt{Kaggle} except \texttt{adform} which is publicly available at \texttt{Harvard Dataverse}.\footnote{\url{http://kaggle.com} and \url{https://dataverse.harvard.edu/}, 2021.} For brevity, we only show 
results using a subset of the datasets, and the remainder of the 
figures are available in \cref{appx:add_figures}.

\begin{table}[!ht]
  \caption{For our study, we consider large tabular datasets with a raw size of approximately 10 GB, whose sizes after feature preprocessing are noted below. The number of rows are represented by $N$, and the number of raw features by $D$.}
  \small
  \label{tab:datasets}
  \centering
  \begin{tabular}{llllll}
    \toprule
	Dataset     & Kind & $N$ & $D$ & GB \\
    \midrule
    adform \citep{DVN/TADBY7_2017} & Binary Classification & 23,999,936  & 108 & 56.2 \\
    adfraud \citep{adfraud2018} & Binary Classification & 149,813,196  & 9 & 30.2 \\
    lendingc \citep{lendingc2019} & Binary Classification & 1,760,668  & 990 & 29.3 \\
    codes \citep{codes2017} & 10-Way Classification & 22,889,691  & 9 & 15.9 \\
    taxifare \citep{taxifare2018} & Regression & 44,936,324  & 17 & 69 \\
    reddit-score \citep{redditscore2019} & Regression & 36,008,714 & 18 & 56.6 \\
    census-income \citep{censusincome2017} & Regression & 2,452,939 & 789 & 38.2 \\
    \bottomrule
  \end{tabular}
\end{table}

\paragraph{Evaluation} For all regression datasets, we maximize the $R^2$ 
score. For all binary classification datasets, we use the 
weighted AUC score, and for multiclass classification datasets, we use
the one-vs-rest formulation of weighted AUC score 
\citep{Fawcett2006AnIT}. We use the implementation provided by
\citet{scikit-learn}. All evaluation scores need to be maximized by 
the HPO algorithm.

\paragraph{HPO Tuning} For our study, we focus on multi-fidelity HPO 
with Hyperband (HB) \citep{Li2017HyperbandAN} and BOHB 
\citep{Falkner2018BOHBRA}. All results are compared to an exhaustive
randomized grid search as the gold standard, where we run each 
algorithm for a total budget of approximately $60000$ seconds 
($\sim 17$ hours) on AWS Sagemaker \citep{PDas2020, Perrone2021AmazonSA} using
\texttt{m5.12/24xlarge} CPU instances. As noted earlier, we use the 
fraction of the full dataset size as the fidelity parameter $r$, which 
is chosen from the set 
$\mathcal{R} = \{ 1/100 , 1/10, 1/4, 1/2, 3/4, 1 \}$. This choice is 
of practical consequence as we describe in \cref{sec:runtime}. \cref{tab:hypers} provides the details of the tuned hyperparameters.\footnote{See \url{https://xgboost.readthedocs.io/en/latest/parameter.html#general-parameters} for the XGBoost hyperparameters.}

\begin{table}[!ht]
  \caption{The set of XGBoost hyperparameters tuned are in the table below, with their considered ranges. For reference, the corresponding XGBoost hyperparameter names are provided alongside the sampling distribution used to sample the range.}
  \label{tab:hypers}
  \centering
  \begin{tabular}{lll}
    \toprule
	Hyperparameter     & XGBoost Parameter & Distribution (Range) \\
    \midrule
    Learning Rate & \texttt{eta} & $\log\text{-}\mathrm{uniform}(10^{-3}, 1.)$ \\
    $\ell_1$ Regularization & \texttt{alpha} & $\log\text{-}\mathrm{uniform}(10^{-6}, 2.)$ \\
    $\ell_2$ Regularization & \texttt{lambda} & $\log\text{-}\mathrm{uniform}(10^{-6}, 2.)$ \\
	Min. Split Loss & \texttt{gamma} & $\log\text{-}\mathrm{uniform}(10^{-6}, 64.)$ \\
    Row Subsample Ratio & \texttt{subsample} & $\mathrm{uniform}(0.5, 1.)$ \\
    Column Subsample Ratio & \texttt{col\_subsample} & $\mathrm{uniform}(0.3, 1.)$ \\
    Max. Tree Depth & \texttt{max\_depth} & $\log\text{-}\mathrm{randint}(2, 8)$ \\
    Boosting Rounds & \texttt{num\_round} & $\log\text{-}\mathrm{randint}(2, 1024)$ \\
    \bottomrule
  \end{tabular}
\end{table}

\subsection{Data Subsampling and Training Runtime}
\label{sec:runtime}

\begin{figure}[!ht]
\centering
\begin{tabular}{cc}
\includegraphics[width=.48\linewidth]{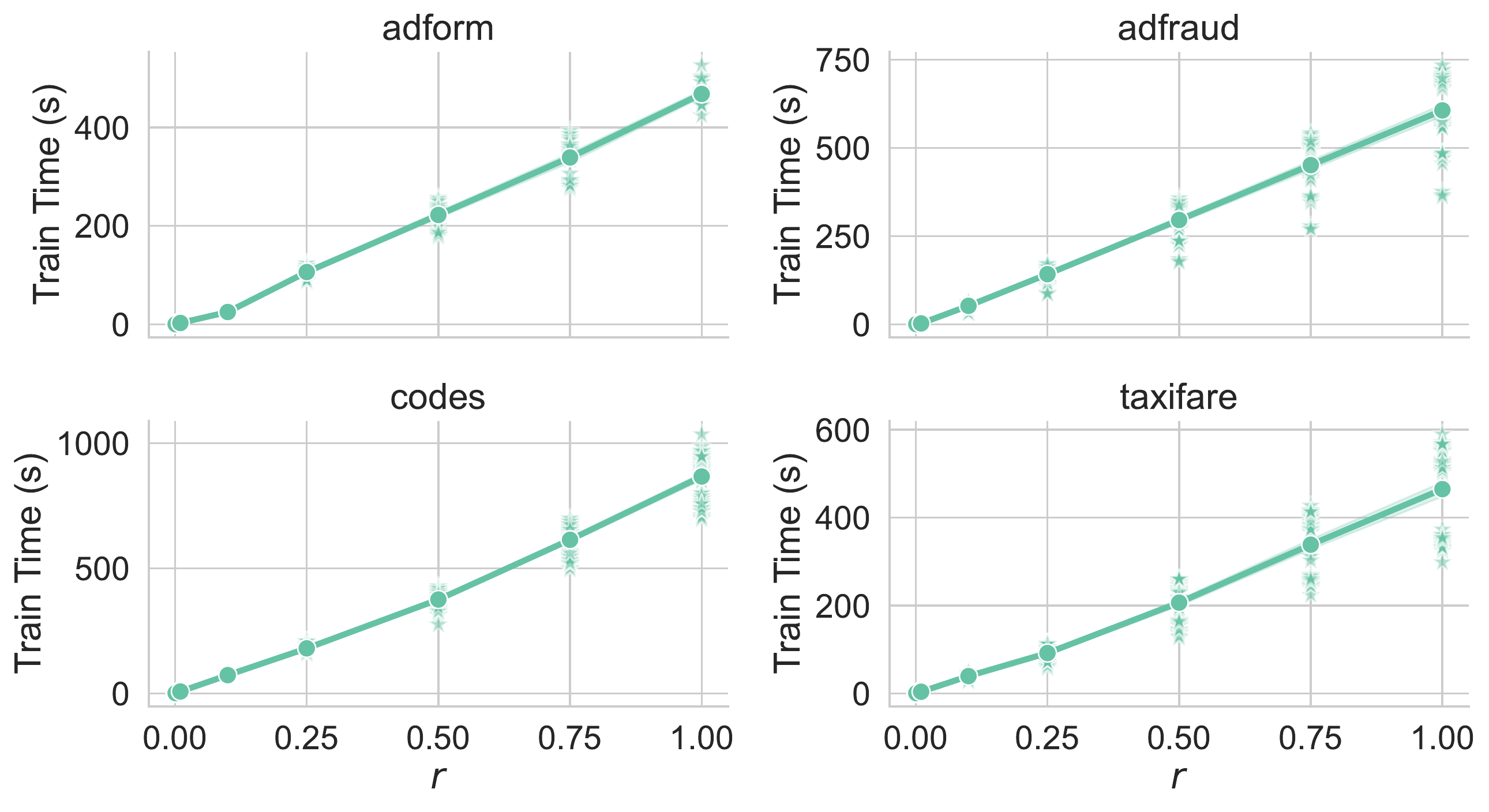} & \includegraphics[width=.48\linewidth]{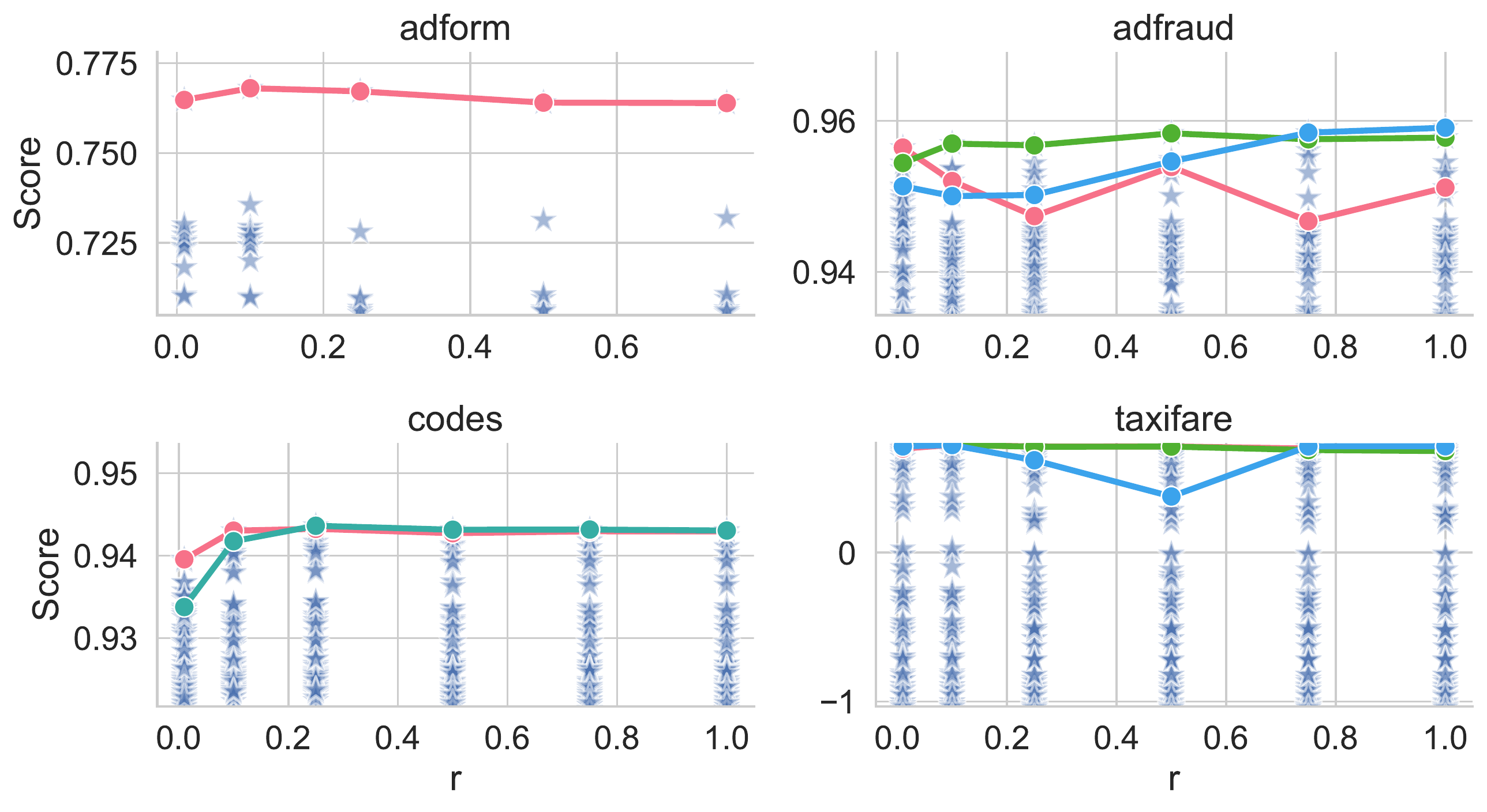}
\end{tabular}
\caption{In these plots, $\star$'s denote individual runs corresponding to different hyperparameter configurations. \textbf{(Left)} We find a linear relationship between XGBoost training time and the dataset size fraction $r \in \mathcal{R}$. This has practical consequences (\cref{sec:runtime}). \textbf{(Right)} XGBoost models satisfy the \emph{ranking hypothesis} \citep{Bornschein2020SmallDB}, making them amenable to multi-fidelity HPO (\cref{sec:rank_hypothesis}). For each dataset, we pick the best performing configuration at each fraction $r$, and see how it performs across all other fractions, connected via a line of the same color. Well-performing configurations in lower-fidelity models typically maintain performance on the full-fidelity model too. We crop the bottom quantile for better legibility. For instance, on the dataset \texttt{adform}, there is only a single line, indicating that the best performing configuration through all fidelities $r$ is the same.}
\label{fig:time_and_rank}
\end{figure}

We find that the relationship between the training 
time of a single XGBoost model and the dataset size is 
roughly linear. This observation is consistent
across all our benchmark datasets when we consider the fraction of
the dataset size $r \in \mathcal{R}$, as visualized in 
\cref{fig:time_and_rank}. Reducing the fraction $r$ further to 1/1,000 
or 1/10,000 does not provide proportional gains to be meaningful in 
practice (often amounting to less than $500\mathrm{ms}$ per run).

This observation has two important practical consequences: (i) the
HPO tuning algorithm can now benefit from faster runs of the
lower fidelity models, and a model using 1/10 the data can train 
roughly 10 times faster than the full-fidelity model; (ii) the linear relationship can be successfully exploited by Hyperband for
efficient resource allocation, since the algorithm expects the relationship to
be roughly linear. A large deviation from linearity would break the
assumptions such that lower-fidelity models end up getting
disproportionately larger time than desired, defeating the resource
allocation strategy of Hyperband.

\subsection{The Ranking Hypothesis}
\label{sec:rank_hypothesis}

\citet{Bornschein2020SmallDB} note that overparameterized neural
network architectures seem to maintain their relative ranking in terms 
of generalization, when trained on arbitrarily small subsets of data.
This is termed as the \emph{ranking hypothesis}, and established
empirically. Neural networks are trained using stochastic minibatches 
of $i.i.d$ data, and a priori, it would appear that batch models like 
XGBoost would not satisfy the ranking hypothesis. Surprisingly, 
however, XGBoost models satisfy the ranking hypothesis for all
our benchmarks considered. Observing \cref{fig:time_and_rank} more
closely reveals that the trend may not always be monotonic from the
lowest fidelity to the highest fidelity and configurations may switch 
ranks. Nevertheless, overall we notice that well-performing lower
fidelity runs tend to perform well also with the full dataset.

As a consequence of this property, we can afford to use far fewer
computational resources to discover competitive hyperparameter
configurations. Moreover, this property is necessary when using 
adaptive resource allocation HPO algorithms such as Hyperband 
\citep{Li2017HyperbandAN}. In practice, we find that optimizing using very small 
data subsets (say $r = 10^{-5}$) can lead to over-regularized XGBoost
models, whose configurations do not perform as well when retrained on the full dataset. This highlights that the choice of the minimum 
fidelity level $r_{\min}$ is crucial to a successful multi-fidelity 
HPO. Our experiments in \cref{sec:relative_perf} and the visuals in 
\cref{fig:relative_perf} reveal that $r=1/100$ works for most 
datasets.

\subsection{Relative Performance of Lower-Fidelity Models}
\label{sec:relative_perf}

\begin{figure}[!ht]
\centering
\begin{tabular}{cc}
\includegraphics[width=.48\linewidth]{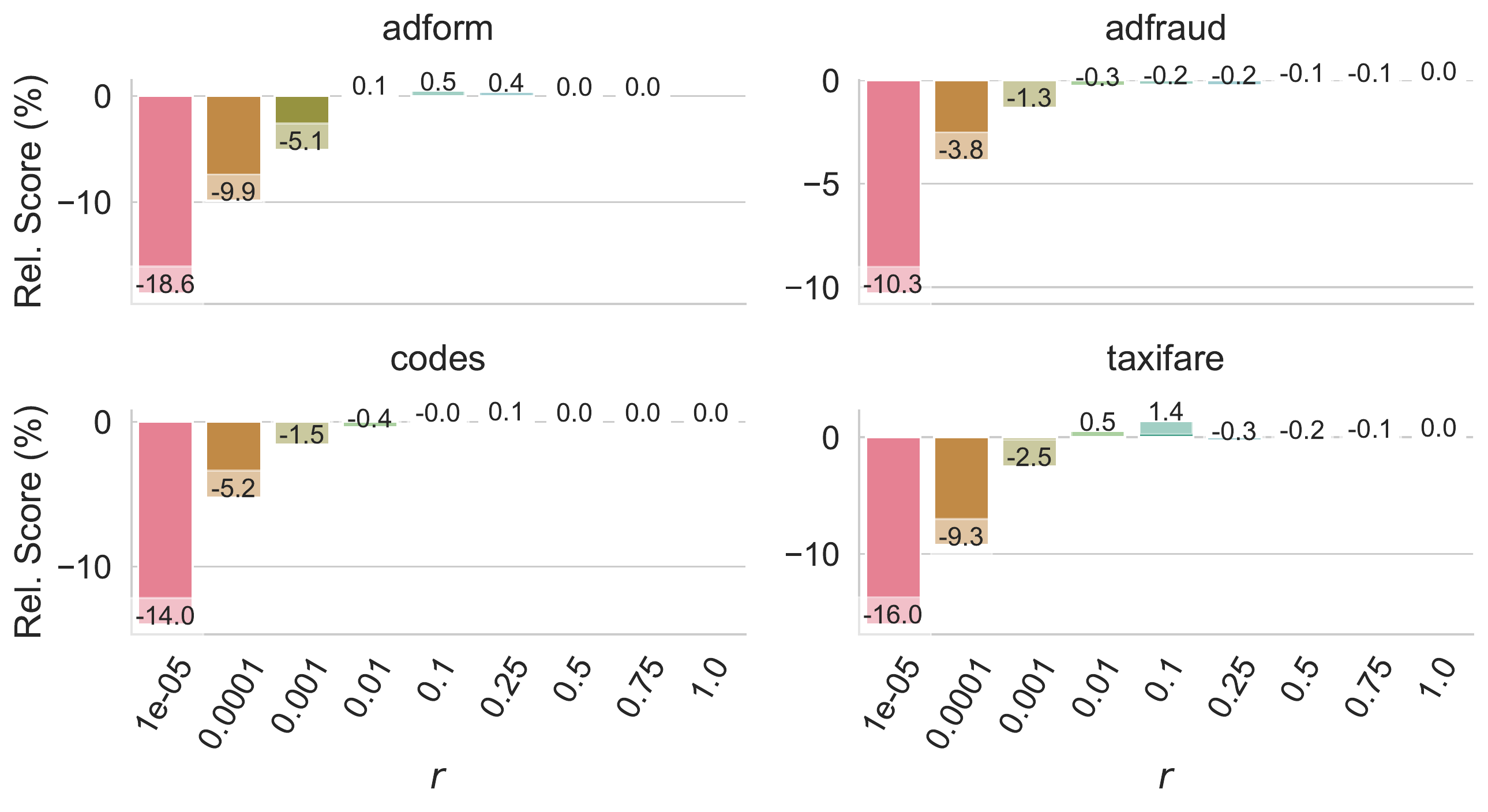} & \includegraphics[width=.48\linewidth]{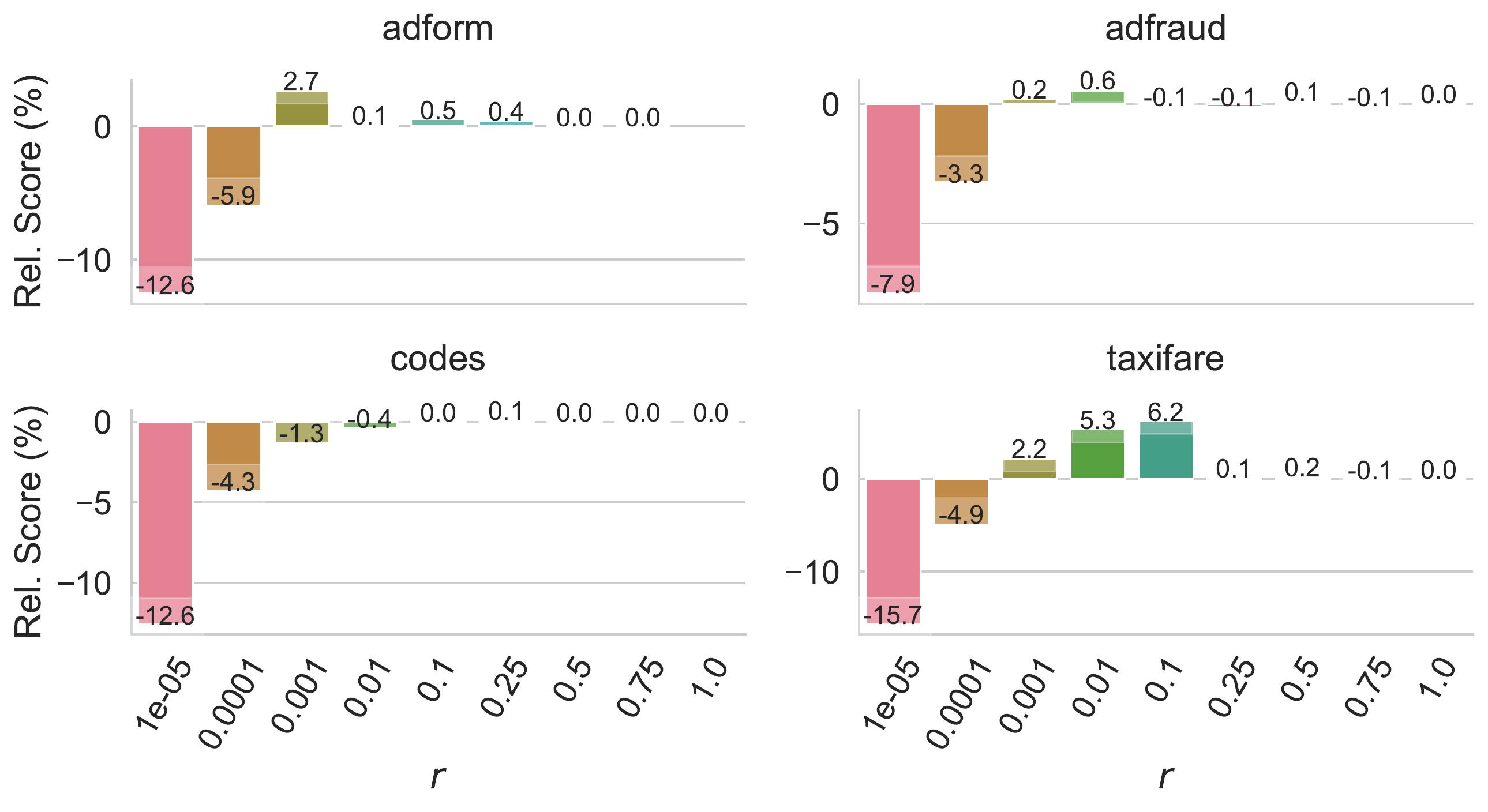}
\end{tabular}
\caption{We compare the relative performance of lower-fidelity XGBoost models to the full-fidelity model trained with the full dataset (\cref{sec:relative_perf}). In addition, to push $r$ to its limit, we include $\{10^{-3},10^{-4},10^{-5}\}$, which suffer a much greater performance drop. \textbf{(Left)} \emph{Without retraining}, lower fidelity models as low as $r=0.01$ (i.e., $1\%$ of the training size) can sustain a reasonably low drop in the validation score. The model effectively breaks when using even smaller subsets. \textbf{(Right)} \emph{With retraining}, we find that the hyperparameter configurations of the lower-fidelity models can further close the generalization gap, often performing better potentially due to the regularization effect of using data subsets.}
\label{fig:relative_perf}
\end{figure}

Considering the best-scoring configuration on the full dataset as the
reference, we quantify the relative performance of lower-fidelity 
models on the validation set in two ways --- (i) \emph{Without 
retraining}: pick best tuned model at each fidelity $r$, 
and directly compute the score. (ii) \emph{With 
retraining}: pick the configuration corresponding to the best tuned 
model at each fidelity $r$, and retrain with the full dataset. To test 
the performance limits of the models, in addition to $\mathcal{R}$, we 
test on $r \in \{10^{-4}, 10^{-5}\}$ as well. The results are 
visualized in \cref{fig:relative_perf}.

In summary, we find that we are able to sustain (on average across 
benchmark datasets) as low as $3.3\%$ drop in performance when 
training with as little as $1\%$ ($r = 1/100$) of the full training 
dataset, sampled uniformly at random. Further, retraining with the
full dataset reduces the generalization gap to just $1.4\%$. Higher 
fidelity models are even better, sustaining less than $0.5\%$ error on
average. This result is of immense practical value as we can
discover competing configurations with far lower computational costs,
as we demonstrate in \cref{sec:eco_hpo}.

\subsection{Economical HPO}
\label{sec:eco_hpo}

By virtue of the facts that, (i) training on data subsets leads to 
proportionately faster training time (\cref{sec:runtime}), (ii) 
XGBoost models satisfy the ranking hypothesis for all practical 
purposes (\cref{sec:rank_hypothesis}), and (iii) lower-fidelity models
can discover high performing configurations \cref{sec:relative_perf},
it is now reasonable to expect benefits in the computational cost of
hyperparameter optimization, especially in terms of the total
wallclock time. 

\begin{figure}[!ht]
\centering
\includegraphics[width=\linewidth]{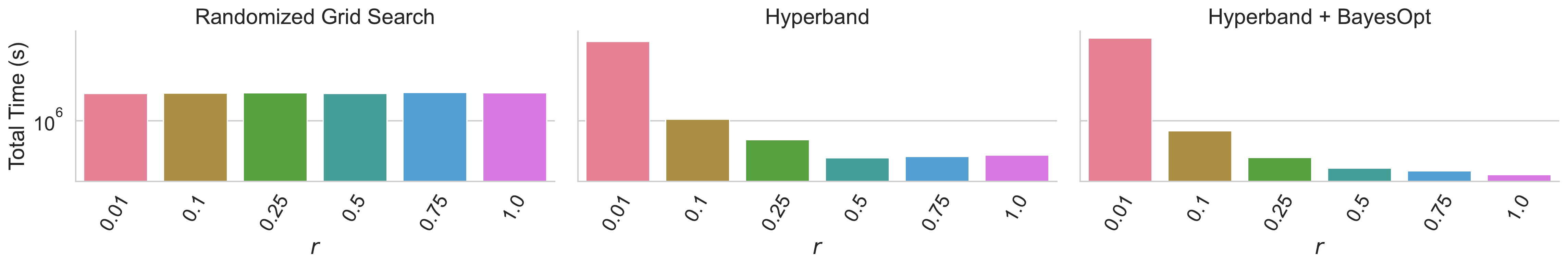}
\caption{Owing to XGBoost models training faster with data subsets, satisfying the ranking hypothesis, and maintaining high-performance with lower-fidelity approximations, we are able to achieve significantly faster wallclock times for the HPO of large-scale XGBoost models, shown here for the \texttt{adform} dataset (\cref{sec:eco_hpo}). The $y$-axis is $\log$-scaled. Notably, Hyperband spends much more time training lower fidelity models, but is able to try many more configurations. Randomized grid search instead spends time in higher-fidelity configurations, which is wasteful if the configuration is not promising.}
\label{fig:eco_hpo}
\end{figure}

To validate this, we compare random search to both our subsampling-based Hyperband proposal and to its BO extension in \cref{fig:eco_hpo}. The results show that higher-fidelity models
now take far less wallclock time, and that we can tune large-scale XGBoost models considerably faster. Further, we find that combining  BO with Hyperband, as in \citet{Falkner2018BOHBRA}, can provide further marginal improvements in the wallclock time of
model tuning. Unlike randomized grid search, which would allocate
roughly the same time to more expensive higher-fidelity configurations,
smarter resource allocation as in Hyperband \citep{Li2017HyperbandAN}
and smarter candidate configuration as with BO
\citep{Falkner2018BOHBRA} can provide meaningful computational cost 
savings.

\section{Conclusions \& Future Outlook}

XGBoost remains an effective model choice for many practical problems 
in the industry, but catering to very large datasets is a 
computational challenge for such batch models, i.e., models which do 
not employ minibatching of the data for stochastic optimization.
Further, small changes in XGBoost hyperparameters can have large effects; for 
instance, changing the tree depth can drastically change the learned
predictor, which one could expect to be a consequence of data 
subsampling.

Our work instead provides surprising evidence to the contrary --- 
XGBoost satisfies many of the favorable properties that allow us to 
exploit multi-fidelity hyperparameter optimization towards faster 
tuning, most importantly the ability to discover promising 
hyperparameter configurations with subsets of data as small as $1\%$ 
of the total size, constructed simply by uniform sampling. 

\paragraph{Limitations \& Future Work} The simplicity and speed of
uniform sampling of the dataset is the key strength of our proposed 
baseline for multi-fidelity hyperparameter optimization. While this 
may be enough for curated datasets, it also remains fundamentally 
limited in its ability to always provide a representative subset for any 
dataset in the wild. Therefore, much of our future effort lies in 
finding reliable ways to summarize datasets using informative 
samples.

\paragraph{Societal Impact} By using uniform 
subsampling to construct data subsets, the results presented in this 
work rely on the often commonly used assumption in machine learning 
that data is $i.i.d.$ and covers the true underlying data distribution 
reasonably well. If the dataset has unfavorable biases towards certain
subpopulations, those may be exacerbated by simple uniform 
subsampling. Better subsampling methods accounting for such scenarios
must be a consideration for practical usage of our proposed baseline 
when such assumptions are violated.

\begin{ack}
We would like to thank Aaron Klein, David Salinas, Giovanni Zappella, Matthias Seeger, C\'edric Archambeau, and Ondrej Bohdal for fruitful discussions.
\end{ack}

\bibliography{references}

\begin{thebibliography}{31}
\providecommand{\natexlab}[1]{#1}
\providecommand{\url}[1]{\texttt{#1}}
\expandafter\ifx\csname urlstyle\endcsname\relax
  \providecommand{\doi}[1]{doi: #1}\else
  \providecommand{\doi}{doi: \begingroup \urlstyle{rm}\Url}\fi

\bibitem[Baeza-Yates and Ribeiro-Neto(2011)]{BaezaYates2011ModernIR}
R.~Baeza-Yates and B.~Ribeiro-Neto.
\newblock {Modern Information Retrieval - the concepts and technology behind
  search, Second edition}.
\newblock 2011.

\bibitem[Bornschein et~al.(2020)Bornschein, Visin, and
  Osindero]{Bornschein2020SmallDB}
J.~Bornschein, Francesco Visin, and Simon Osindero.
\newblock {Small Data, Big Decisions: Model Selection in the Small-Data
  Regime}.
\newblock In \emph{ICML}, 2020.

\bibitem[Bottou(2012)]{Bottou2012StochasticGD}
L.~Bottou.
\newblock {Stochastic Gradient Descent Tricks}.
\newblock In \emph{Neural Networks: Tricks of the Trade}, 2012.

\bibitem[Bureau(2017)]{censusincome2017}
US~Census Bureau.
\newblock {2013 American Community Survey}, May 2017.
\newblock URL
  \url{https://www.kaggle.com/census/2013-american-community-survey}.

\bibitem[Chen and Guestrin(2016)]{Chen2016XGBoostAS}
Tianqi Chen and Carlos Guestrin.
\newblock {XGBoost: A Scalable Tree Boosting System}.
\newblock \emph{Proceedings of the 22nd ACM SIGKDD International Conference on
  Knowledge Discovery and Data Mining}, 2016.

\bibitem[Cloud(2018)]{taxifare2018}
Google Cloud.
\newblock {New York City Taxi Fare Prediction}, Sep 2018.
\newblock URL
  \url{https://www.kaggle.com/c/new-york-city-taxi-fare-prediction}.

\bibitem[Das et~al.(2020)Das, Perrone, Ivkin, Bansal, Karnin, Shen,
  Shcherbatyi, Elor, Wu, Zolic, Lienart, Tang, Ahmed, Faddoul, Jenatton,
  Winkelmolen, Gautier, Dirac, Perunicic, Miladinovic, Zappella, Archambeau,
  Seeger, Dutt, and Rouesnel]{PDas2020}
Piali Das, Valerio Perrone, Nikita Ivkin, Tanya Bansal, Zohar Karnin, Huibin
  Shen, Iaroslav Shcherbatyi, Yotam Elor, Wilton Wu, Aida Zolic, Thibaut
  Lienart, Alex Tang, Amr Ahmed, Jean~Baptiste Faddoul, Rodolphe Jenatton, Fela
  Winkelmolen, Philip Gautier, Leo Dirac, Andre Perunicic, Miroslav
  Miladinovic, Giovanni Zappella, Cédric Archambeau, Matthias Seeger, Bhaskar
  Dutt, and Laurence Rouesnel.
\newblock {Amazon SageMaker Autopilot: a white box AutoML solution at scale}.
\newblock \emph{Proceedings of the Fourth International Workshop on Data
  Management for End-to-End Machine Learning}, 2020.

\bibitem[Falkner et~al.(2018)Falkner, Klein, and Hutter]{Falkner2018BOHBRA}
S.~Falkner, Aaron Klein, and F.~Hutter.
\newblock {BOHB: Robust and Efficient Hyperparameter Optimization at Scale}.
\newblock In \emph{ICML}, 2018.

\bibitem[Fawcett(2006)]{Fawcett2006AnIT}
Tom Fawcett.
\newblock {An introduction to ROC analysis}.
\newblock \emph{Pattern Recognit. Lett.}, 27:\penalty0 861--874, 2006.

\bibitem[Frazier(2018)]{Frazier2018ATO}
P.~Frazier.
\newblock {A Tutorial on Bayesian Optimization}.
\newblock \emph{ArXiv}, abs/1807.02811, 2018.

\bibitem[George(2019)]{lendingc2019}
Nathan George.
\newblock {All Lending Club loan data}, Apr 2019.
\newblock URL \url{https://www.kaggle.com/wordsforthewise/lending-club}.

\bibitem[Hartmanis and Leeuwen(2002)]{Hartmanis2002NeuralNT}
J.~Hartmanis and J.~V. Leeuwen.
\newblock {Neural Networks: Tricks of the Trade}.
\newblock In \emph{Lecture Notes in Computer Science}, 2002.

\bibitem[He et~al.(2014)He, Pan, Jin, Xu, Liu, Xu, Shi, Atallah, Herbrich,
  Bowers, and Candela]{He2014PracticalLF}
Xinran He, Junfeng Pan, Ou~Jin, T.~Xu, Bo~Liu, Tao Xu, Yanxin Shi, Antoine
  Atallah, R.~Herbrich, S.~Bowers, and J.~Q. Candela.
\newblock {Practical Lessons from Predicting Clicks on Ads at Facebook}.
\newblock In \emph{ADKDD'14}, 2014.

\bibitem[Jamieson and Talwalkar(2016)]{Jamieson2016NonstochasticBA}
Kevin~G. Jamieson and Ameet~S. Talwalkar.
\newblock {Non-stochastic Best Arm Identification and Hyperparameter
  Optimization}.
\newblock \emph{ArXiv}, abs/1502.07943, 2016.

\bibitem[Kadra et~al.(2021)Kadra, Lindauer, Hutter, and
  Grabocka]{Kadra2021RegularizationIA}
Arlind Kadra, M.~Lindauer, F.~Hutter, and Josif Grabocka.
\newblock {Regularization is all you Need: Simple Neural Nets can Excel on
  Tabular Data}.
\newblock \emph{ArXiv}, abs/2106.11189, 2021.

\bibitem[Kaggle(2019)]{redditscore2019}
Kaggle.
\newblock {May 2015 Reddit Comments}, Jun 2019.
\newblock URL \url{https://www.kaggle.com/kaggle/reddit-comments-may-2015}.

\bibitem[Klein et~al.(2017)Klein, Falkner, Bartels, Hennig, and
  Hutter]{Klein2017FastBO}
Aaron Klein, S.~Falkner, Simon Bartels, Philipp Hennig, and F.~Hutter.
\newblock {Fast Bayesian Optimization of Machine Learning Hyperparameters on
  Large Datasets}.
\newblock \emph{ArXiv}, abs/1605.07079, 2017.

\bibitem[Li et~al.(2017)Li, Jamieson, DeSalvo, Rostamizadeh, and
  Talwalkar]{Li2017HyperbandAN}
Lisha Li, Kevin~G. Jamieson, Giulia DeSalvo, Afshin Rostamizadeh, and Ameet~S.
  Talwalkar.
\newblock {Hyperband: A Novel Bandit-Based Approach to Hyperparameter
  Optimization}.
\newblock \emph{J. Mach. Learn. Res.}, 18:\penalty0 185:1--185:52, 2017.

\bibitem[Mockus(1977)]{Mockus1977OnBM}
J.~Mockus.
\newblock {On Bayesian Methods for Seeking the Extremum and their Application}.
\newblock In \emph{IFIP Congress}, 1977.

\bibitem[Nickson et~al.(2014)Nickson, Osborne, Reece, and
  Roberts]{Nickson2014AutomatedML}
T.~Nickson, Michael~A. Osborne, S.~Reece, and S.~Roberts.
\newblock {Automated Machine Learning using Stochastic Algorithm Tuning}.
\newblock 2014.

\bibitem[Pedregosa et~al.(2011)Pedregosa, Varoquaux, Gramfort, Michel, Thirion,
  Grisel, Blondel, Prettenhofer, Weiss, Dubourg, Vanderplas, Passos,
  Cournapeau, Brucher, Perrot, and Duchesnay]{scikit-learn}
F.~Pedregosa, G.~Varoquaux, A.~Gramfort, V.~Michel, B.~Thirion, O.~Grisel,
  M.~Blondel, P.~Prettenhofer, R.~Weiss, V.~Dubourg, J.~Vanderplas, A.~Passos,
  D.~Cournapeau, M.~Brucher, M.~Perrot, and E.~Duchesnay.
\newblock {Scikit-learn: Machine Learning in {P}ython}.
\newblock \emph{Journal of Machine Learning Research}, 12:\penalty0 2825--2830,
  2011.

\bibitem[Perrone et~al.(2021)Perrone, Shen, Zolic, Shcherbatyi, Ahmed, Bansal,
  Donini, Winkelmolen, Jenatton, Faddoul, Pogorzelska, Miladinovic, Kenthapadi,
  Seeger, and Archambeau]{Perrone2021AmazonSA}
Valerio Perrone, Huibin Shen, Aida Zolic, I.~Shcherbatyi, A.~Ahmed, Tanya
  Bansal, Michele Donini, Fela Winkelmolen, Rodolphe Jenatton, J.~Faddoul,
  Barbara Pogorzelska, Miroslav Miladinovic, K.~Kenthapadi, M.~Seeger, and
  C.~Archambeau.
\newblock {Amazon SageMaker Automatic Model Tuning: Scalable Gradient-Free
  Optimization}.
\newblock \emph{Proceedings of the 27th ACM SIGKDD Conference on Knowledge
  Discovery \& Data Mining}, 2021.

\bibitem[Rasmussen and Williams(2005)]{Rasmussen2005GaussianPF}
C.~Rasmussen and Christopher K.~I. Williams.
\newblock {Gaussian Processes for Machine Learning (Adaptive Computation and
  Machine Learning)}.
\newblock 2005.

\bibitem[Shahriari et~al.(2016)Shahriari, Swersky, Wang, Adams, and
  de~Freitas]{7352306}
Bobak Shahriari, Kevin Swersky, Ziyu Wang, Ryan~P. Adams, and Nando de~Freitas.
\newblock {Taking the Human Out of the Loop: A Review of Bayesian
  Optimization}.
\newblock \emph{Proceedings of the IEEE}, 104\penalty0 (1):\penalty0 148--175,
  2016.
\newblock \doi{10.1109/JPROC.2015.2494218}.

\bibitem[Shioji(2017)]{DVN/TADBY7_2017}
Enno Shioji.
\newblock {Adform click prediction dataset}, 2017.
\newblock URL \url{https://doi.org/10.7910/DVN/TADBY7}.

\bibitem[Shwartz-Ziv and Armon(2021)]{ShwartzZiv2021TabularDD}
Ravid Shwartz-Ziv and A.~Armon.
\newblock {Tabular Data: Deep Learning is Not All You Need}.
\newblock \emph{ArXiv}, abs/2106.03253, 2021.

\bibitem[Somepalli et~al.(2021)Somepalli, Goldblum, Schwarzschild, Bruss, and
  Goldstein]{Somepalli2021SAINTIN}
G.~Somepalli, Micah Goldblum, Avi Schwarzschild, C.~B. Bruss, and T.~Goldstein.
\newblock {SAINT: Improved Neural Networks for Tabular Data via Row Attention
  and Contrastive Pre-Training}.
\newblock \emph{ArXiv}, abs/2106.01342, 2021.

\bibitem[Swersky et~al.(2013)Swersky, Snoek, and Adams]{Swersky2013MultiTaskBO}
Kevin Swersky, Jasper Snoek, and Ryan~P. Adams.
\newblock {Multi-Task Bayesian Optimization}.
\newblock In \emph{NIPS}, 2013.

\bibitem[TalkingData(2018)]{adfraud2018}
TalkingData.
\newblock {AdTracking Fraud Detection Challenge}, May 2018.
\newblock URL
  \url{https://www.kaggle.com/c/talkingdata-adtracking-fraud-detection/overview/description}.

\bibitem[Turner et~al.(2020)Turner, Eriksson, McCourt, Kiili, Laaksonen, Xu,
  and Guyon]{Turner2020BayesianOI}
Ryan Turner, David Eriksson, M.~McCourt, Juha Kiili, Eero Laaksonen, Zhen Xu,
  and I.~Guyon.
\newblock {Bayesian Optimization is Superior to Random Search for Machine
  Learning Hyperparameter Tuning: Analysis of the Black-Box Optimization
  Challenge 2020}.
\newblock In \emph{NeurIPS}, 2020.

\bibitem[Zavadskyy(2017)]{codes2017}
Vladislav Zavadskyy.
\newblock {Lots of code}, Dec 2017.
\newblock URL \url{https://www.kaggle.com/zavadskyy/lots-of-code}.

\end{thebibliography}
\bibliographystyle{plainnat}

\clearpage

\clearpage
\appendix

\section{Additional Figures}
\label{appx:add_figures}

\subsection{Data Subsampling and Training Runtime}

\begin{figure}[!ht]
\centering
\begin{tabular}{c}
\includegraphics[width=\linewidth]{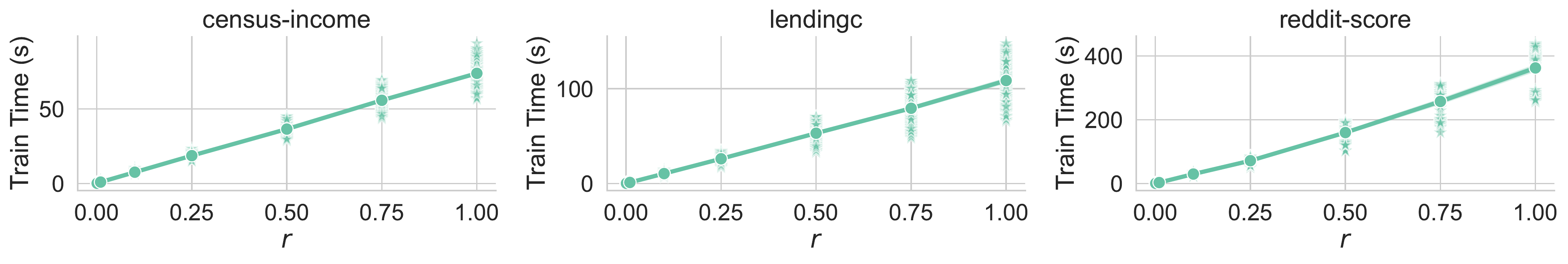}
\end{tabular}
\caption{As in \cref{fig:time_and_rank}(a), for the remainder of our 
benchmark datasets too, we find a strong linear relationship.}
\label{fig:time_appendix}
\end{figure}

In continuation to \cref{sec:runtime}, we provide the training runtime
plots for the remainder of our benchmark datasets (see 
\cref{tab:datasets}) in \cref{fig:time_appendix}.

\subsection{The Ranking Hypothesis}

\begin{figure}[!ht]
\centering
\begin{tabular}{c}
\includegraphics[width=\linewidth]{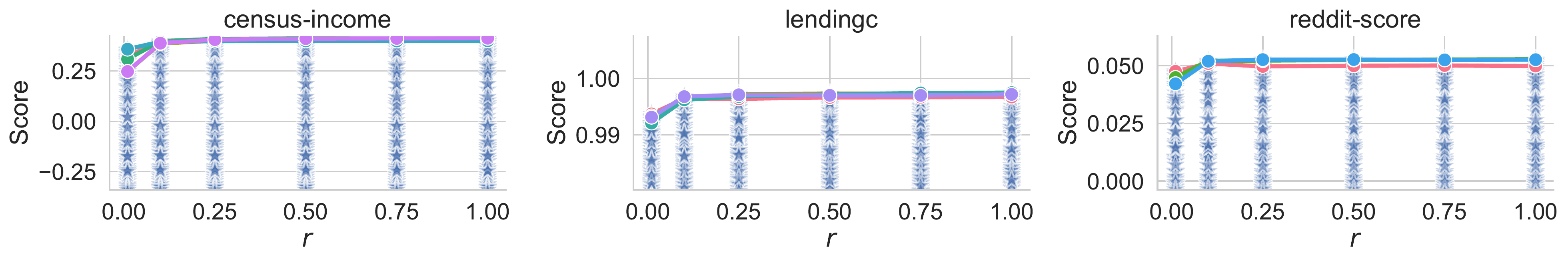}
\end{tabular}
\caption{As in \cref{fig:time_and_rank}(b), the remainder of our 
benchmark datasets satisfy \emph{the ranking hypothesis} as well.}
\label{fig:rank_appendix}
\end{figure}

For the remainder of the datasets of our benchmark in 
\cref{fig:rank_appendix}, we are able to demonstrate \emph{the ranking 
hypothesis} is satisfied. The consequences of this are discussed in 
\cref{sec:rank_hypothesis}.

\subsection{Relative Performance of Lower-Fidelity Models}

For the remainder of the datasets, we make a similar assessment 
\emph{without retraining} and \emph{with retraining}, as described in 
\cref{sec:relative_perf}.

\begin{figure}[!ht]
\centering
\begin{tabular}{c}
\includegraphics[width=\linewidth]{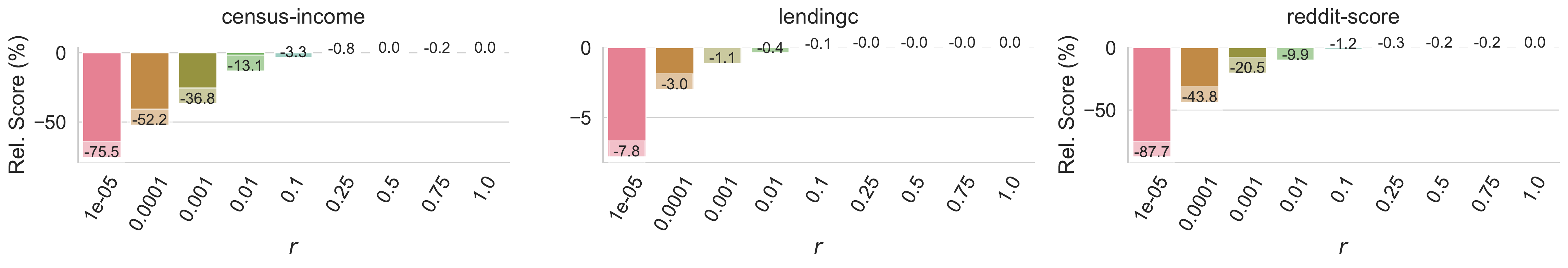} \\
(a) \emph{without retraining} \\
\includegraphics[width=\linewidth]{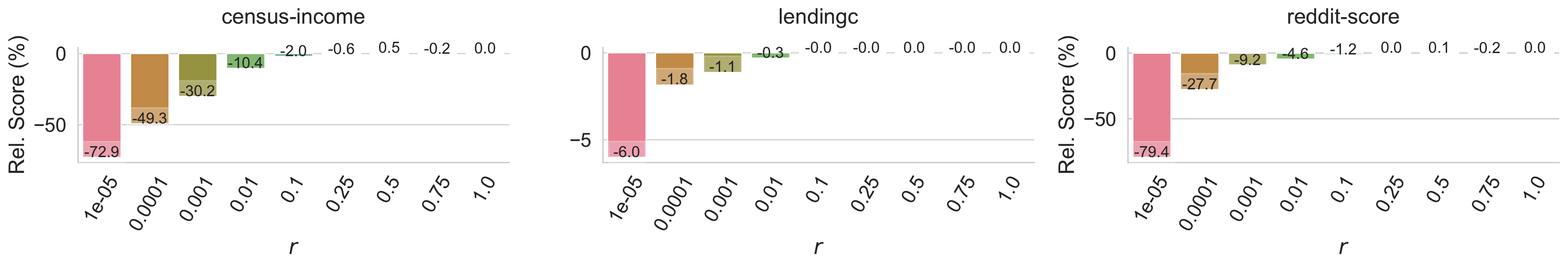} \\
(b) \emph{with retraining} \\
\end{tabular}
\caption{As in \cref{fig:relative_perf}, the remainder of our 
benchmark datasets also show similar trends in performance with 
uniformly subsampled datasets. Here again, we show much lower values 
of $r$ to push the subsampling to its limits.}
\label{fig:rel_perf_appendix}
\end{figure}

\subsection{Economical HPO}

\begin{figure}[!ht]
\centering
\begin{tabular}{c}
\includegraphics[width=\linewidth]{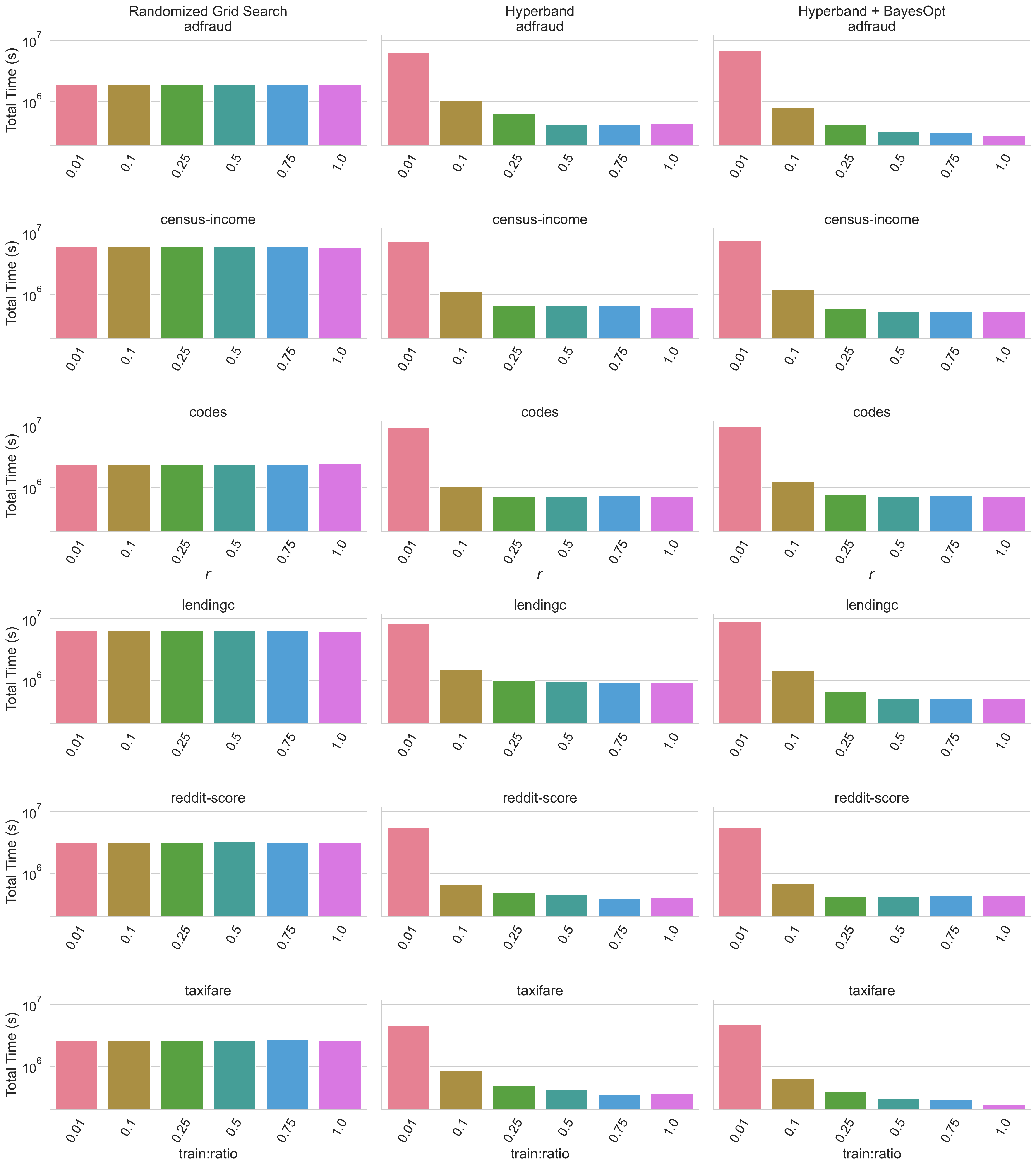}
\end{tabular}
\caption{As in \cref{fig:eco_hpo}, the remainder of our 
benchmark datasets reveal similar runtime trends, where combining with 
Bayesian optimization can often have practical benefits. More 
importantly, by virtue of the Hyperband resource scheduling, we are 
able to test many more configurations and only spend higher resources 
on the most promising ones.}
\label{fig:econ_hpo_appendix}
\end{figure}

We provide the cumulative tuning time plots, as in \cref{sec:eco_hpo},
for the remainder of the datasets in \cref{fig:econ_hpo_appendix}.

\end{document}